\def\year{2020}\relax
\documentclass[letterpaper]{article} 
\usepackage{aaai20}  
\usepackage{times}  
\usepackage{helvet} 
\usepackage{courier}  
\usepackage[hyphens]{url}  
\usepackage{graphicx} 
\usepackage{graphicx}  
\usepackage{multirow}
\usepackage{amssymb}

\title{Task-Oriented Dialog Systems that Consider Multiple Appropriate Responses under the Same Context}

\author{Yichi Zhang\thanks{This work was done during Yichi Zhang's summer  internship at University of California, Davis.}\\
	Tsinghua University \\
	Beijing, China  \\
	{zhangyic17@mails.tsinghua.edu.cn} \\\And
	Zhijian Ou \\
	Tsinghua University \\
	Beijing, China\\
	{ozj@tsinghua.edu.cn} \\\And
	Zhou Yu \\
	University of California, Davis \\
	Davis, California, USA \\
	{joyu@ucdavis.edu} \\}

\begin{document}
	
	\maketitle
	
	\begin{abstract}
		Conversations have an intrinsic one-to-many property, which means that multiple responses can be appropriate for the same dialog context. In task-oriented dialogs, this property leads to different valid dialog policies towards task completion. However, none of the existing task-oriented dialog generation approaches takes this property into account. 
		We propose a Multi-Action Data Augmentation (MADA) framework to utilize the one-to-many property to generate diverse appropriate dialog responses. 
		Specifically, we first use dialog states to summarize the dialog history, and then discover all possible mappings from every dialog state to its different valid system actions.
		During dialog system training, we enable the current dialog state to map to all valid system actions discovered in the previous process to create additional state-action pairs.
		By incorporating these additional pairs, the dialog policy learns a balanced action distribution, which further guides the dialog model to generate diverse responses. 
		Experimental results show that the proposed framework consistently improves dialog policy diversity, and results in improved response diversity and appropriateness. Our model obtains  state-of-the-art results on MultiWOZ. 
		
	\end{abstract}
	
	\section{Introduction}
	\label{intro}
	One big challenge in dialog system generation is that multiple responses can be appropriate under the same conversation context. This challenge originated from the intrinsic diversity of human conversations.
	Although recent progress in sequence-to-sequence (seq2seq) learning \cite{sutskever2014sequence} improves dialog systems performance \cite{serban2017hierarchical,wen2017network,lei2018sequicity}. These systems still ignore this one-to-many property in conversation. Therefore, they are not able to handle diverse user behaviors in real-world settings \cite{li2016diversity,rajendran2018learning}.

	\begin{figure}[t]
		\centering
		\includegraphics[width=1.0\columnwidth]{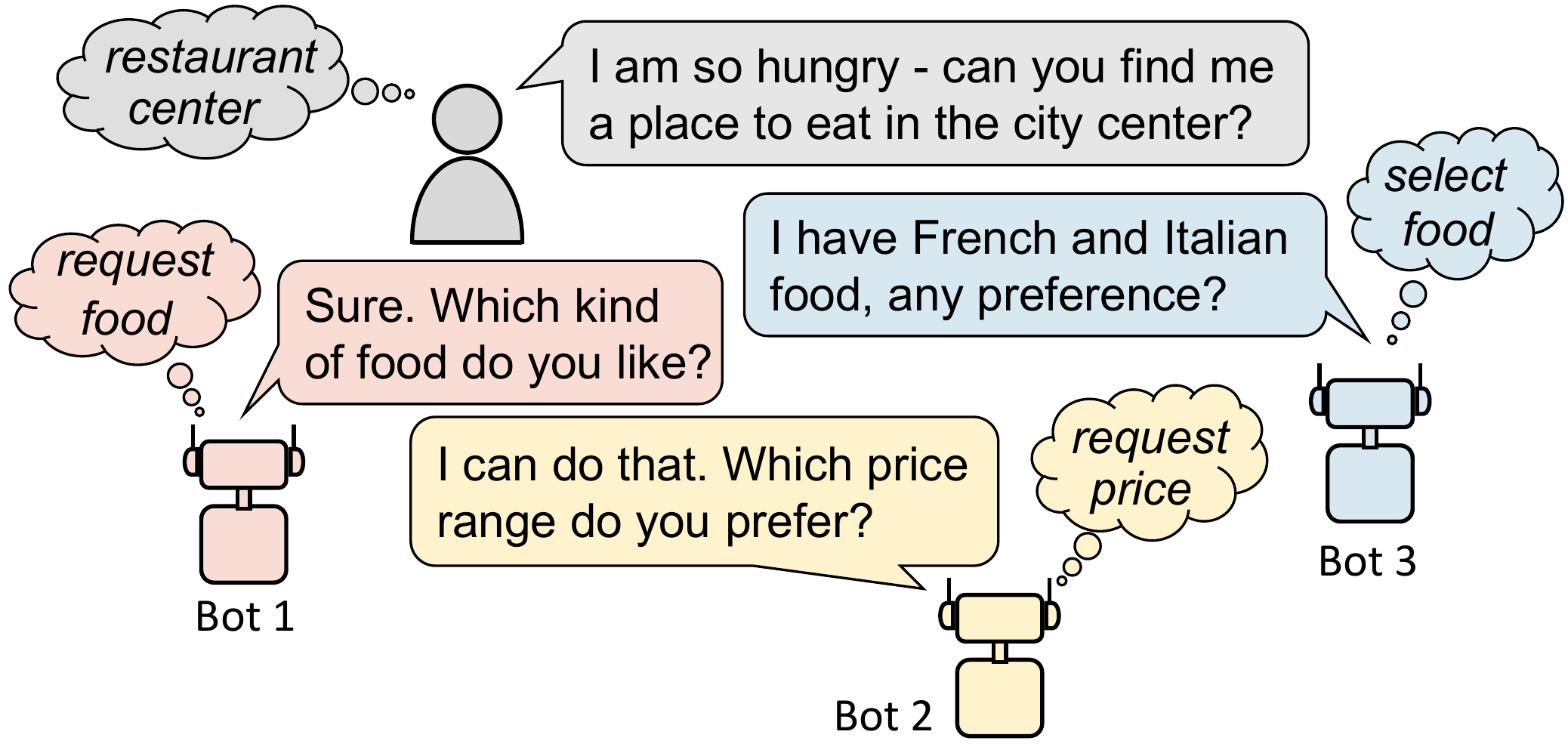} 
		\caption{Multiple responses produced by different dialog policies (shown in clouds) are proper for the same context. }
		\label{example}
	\end{figure}

	Previous studies model this one-to-many conversation property to improve \textit{utterance-level} diversity in open-domain dialog generation \cite{zhao2017learning,zhou2017mechanism,zhou2018elastic}. 
	None of previous task-oriented systems consider such one-to-many property, since they focus on task completion policies instead of language variations. However, the one-to-many phenomenon is also prevalent in task-oriented dialogs, in the form of different responding policies for the same dialog context (Fig.\ref{example}).
	Since in collected dialog datasets each dialog context has only one reference response, the distribution of valid system actions for each dialog state rely on their occurring frequencies in the datasets which are usually highly unbalanced.
	Models trained on these unbalanced datasets tend to capture the most common dialog policy but ignore rarely occurred yet feasible user behaviors, which results in learning skewed and low-coverage policies.
	
	Our goal is to address such data bias and model this one-to-many property in task-oriented dialogs to enrich dialog policy diversity, therefore building dialog systems that can generate more diverse system responses. Instead of simply learning how to map one user response to many system responses \cite{rajendran2018learning}
	, we propose to discover the mapping from one dialog state (condensed dialog history) to multiple system actions and then generate system responses conditioned on learned actions. Since the number of unique dialog states and system actions are much smaller than the number of unique user and system responses, the mapping is more structured and easier to incorporate in learning. 
	
	Specifically, we propose a general Multi-Action Data Augmentation (MADA) framework to achieve such mapping. 
	We first delexicalize all utterances to reduce surface language diversity. 
	Then we use dialog states and system actions to achieve condensed but sufficient information representation.
	We accumulate all mappings from dialog states to valid system actions from the entire training corpus.
	Finally in the dialog system training process, we force the model to not only take the ground truth system action as training sample, but also create extra training samples by including other possible system actions that are valid under that dialog state based on the state-action mapping obtained earlier. Then the learned policy is able to produce a more balanced system action distribution given a dialog context. Therefore, the dialog system can produce a set of diverse and valid system actions, which further guide the model to generate diverse and appropriate responses. 
	
	We evaluate the proposed method on MultiWOZ \cite{budzianowski2018multiwoz}, a large-scale human-human task-oriented dialog dataset covering multiple domains. We show that our data augmentation method significantly improves response generation quality on various learning models. To utilize the one-to-many mapping under the challenging multi-domain scenario, we propose Domain Aware Multi-Decoder (DAMD) network to accommodate state-action pair structure in generation.  Our model obtains the new state-of-the-art results on MultiWOZ's response generation task. Human evaluation results show that DAMD with data augmentation generates diverse and valid responses. 
	

	\section{Related Work}
	The trend of building task-oriented systems is changing from training separate dialog models independently \cite{young2013pomdp,wen2015semantically,liu2016attention,mrkvsic2017neural}
	to end-to-end trainable dialog models \cite{zhao2017generative,wen2017network,eric2017key}. 
	Specifically, \citeauthor{lei2018sequicity} \shortcite{lei2018sequicity} propose a two stage seq2seq model (Sequicity) with copy mechanism \cite{gu2016incorporating} that completes the dialog state tracking and response generation jointly via a single seq2seq architecture. 
	These systems achieve promising results, however, all of these models are designed for a specific domain which lacks the generalization ability to multi-domains, e.g. the recently proposed multi-domain dataset MultiWOZ \cite{budzianowski2018multiwoz}. Although several models are proposed to handle the multi-domain response generation task \cite{zhao2019rethinking,mehri2019structured,chen2019semantically}, the generation quality is far from perfect, presumably due to the complex task definitions, large policy coverage and flexible language styles. We believe by modeling the one-to-many dialog property, we can improve multi-domain dialog system generation.
	
	The one-to-many problem is more noticeable in open-domain social dialog systems since ``I don't know" can be valid response to all questions, but such response is not very useful or engaging. Therefore, previous social response generation methods attempt to model the one-to-many property by modeling responses with other meta information, such as response specificity \cite{zhou2017mechanism,zhou2018elastic}. By considering these meta information, the model can generate social dialog response with larger diversity. 
	However, for task-oriented dialog systems, the only work that models this one-to-many property utilizes this property to retrieve dialog system responses instead of generating response \cite{rajendran2018learning}. We propose to take advantage of this one-to-many mapping property to generate more diverse dialog responses in task-oriented dialog systems. Moreover, one key advantage of our proposed framework is that the multiple actions decoded by the dialog model are interpretable and controllable. We leverage different diverse decoding methods \cite{li2016simple,fan2018hierarchical,holtzman2019curious} to improve the diversity of generated system actions. 

	\section{Multi-Action Data Augmentation Framework}
	
\begin{figure}[t!]
		\centering
		\includegraphics[width=1.0\columnwidth]{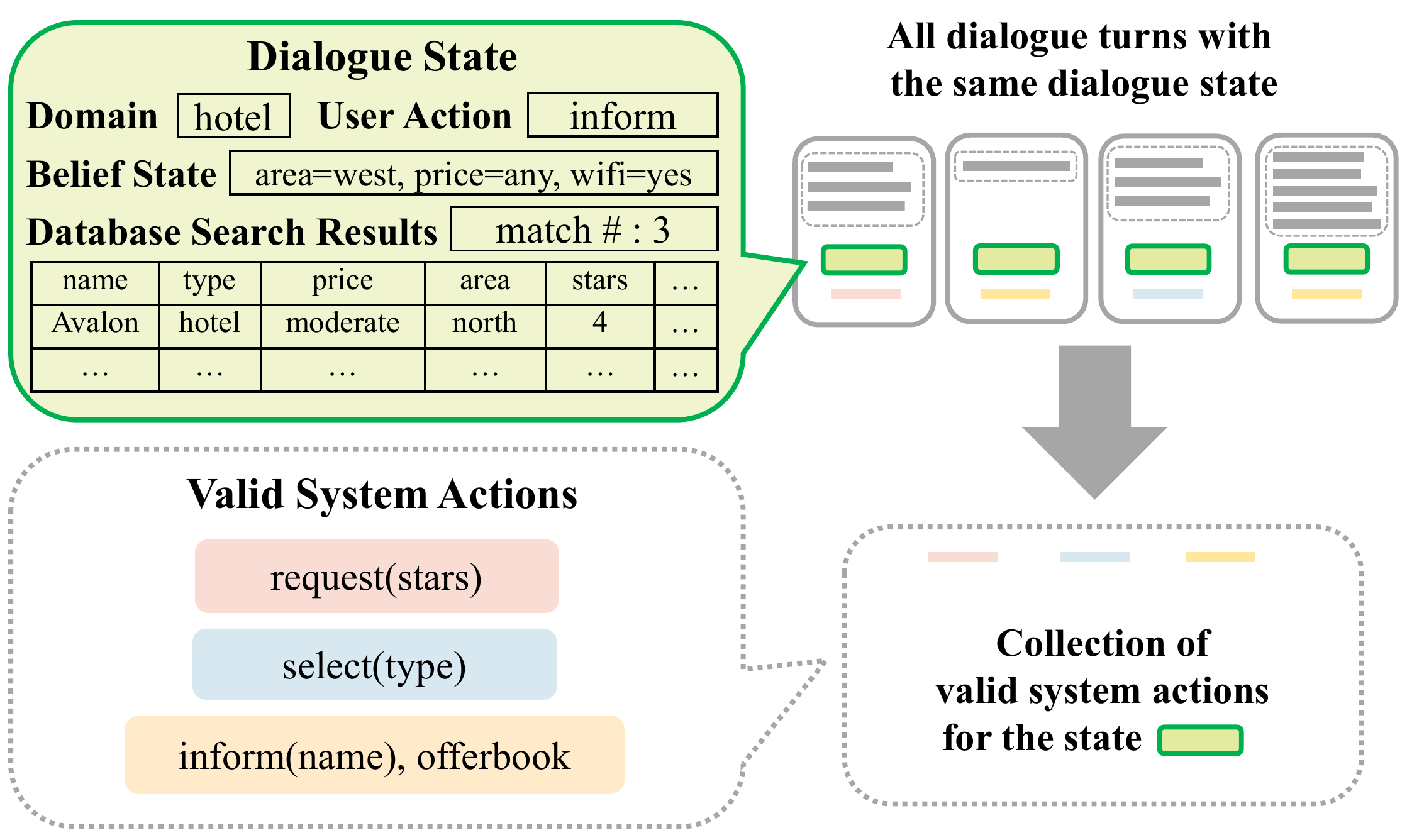} 
		\caption{The overview of our Multi-Action Data Augmentation (MADA) framework. The green blocks denotes the same dialog state, and bars with different colors are different valid system actions corresponding to this state. Other valid state-action pairs are additional training data to learn the state-to-action mapping.}
		\label{framework}
	\end{figure}

	We introduce the Multi-Action Data Augmentation (MADA) framework that is generalizable to all task-oriented dialog scenarios. MADA is suitable to all dialog models that take system action supervision. It aims to increase dialog response generation diversity through learning a dialog policy that decodes a diverse set of valid system actions when given a dialog context.
	In MADA, we first discover the one-to-many mapping of a summarized dialog context (i.e. dialog state) to a set of system actions that are appropriate under that context. We then make the dialog model to include all the additional actions that are valid according to the one state to many system action mapping during training. In this way, the dialog policy is trained by a balanced mapping between dialog state and different system actions. Therefore, in the end the policy can generate a diverse set of system actions that are all appropriate under a give context. Such a diverse set of system actions will naturally lead to diverse system responses. Fig.\ref{framework} shows an example dialog state to multiple actions mapping. 
	
	To learn this one-to-many mapping, we first need to design suitable dialog state and system action that are sufficient to represent dialog policy learning. Dialog state needs to summarize the dialog history that contains sufficient information for a dialog system to decide what actions to take next. So we define dialog state $S_t$ at turn $t$ to have four types of information: 1) current dialog domain, 2) belief state, 3) database search results and 4) current user action. 
	Current dialog domain $D_t$ is essential, because one single task can have multiple dialog domains, so the active domain is necessary to include in the dialog state representation. Belief state $B_t$ is necessary because the belief state records slots and corresponding values informed by user in each turn, e.g. ``\textit{price=cheap, location=west}", these slots are useful in searching database to obtain task information. Database (DB) search results $DB_t$ also influence the next system action, because based on the data search results, the system may request for an unmentioned slot to reduce the search range. Finally current user action $A^U_t$ can also influence the system policy, because sometimes the system need to give direct feedback to the user, such as providing a phone number when it is asked.
	\begin{equation}
	{S}_t\triangleq \langle D_t, B_t, {DB}_t, A^U_t \rangle
	\end{equation}
	
	System action is the semantic representation of the system utterance. We define system action consists of dialog domains, dialog acts, and slots. One example system action is ``\textit{hotel-request(price, area)}". 
	
	We then go through the entire training data to find system response that share the same dialog state to form all the one state to many action mappings.
	Finally, we introduce how to train a dialog policy that produces a balanced valid action distribution under each dialog state.
	
	Training a dialog policy is to learn the optimal mapping from dialog states to system actions towards achieving task goals efficiently. In another way, we are learning the correct dialog actions conditioned on a dialog state:
	\begin{equation}
	\mathcal{L}=\sum_{t\in\mathcal{D}}\log P(A_t|S_t)
	\end{equation}
	
	Due to the one-to-many property, for a specific dialog state $S$, there exists $K$ different system actions $A^{(1)}, \dots, A^{(K)}$ that are valid for this state, i.e. for $i=1,\dots,K, \  \exists t\in\mathcal{D}\ \  {\rm s.t.}\ (S_t, A_t)=(S,A^{(i)}) $, and we denote the valid system action set as $\mathcal{V}(S)$. If some state-action pairs $(S, A^{(j)})$ have much lower frequency than other pairs $(S, A^{(k)})$,
	then the model tends to only capture the majority mappings and ignores the minority ones. So the trained dialog policy lacks diversity. This problem is also known as a general drawback of the maximum likelihood estimate over unbalance dataset \cite{jennrich1986unbalanced}. 
	
	We address this issue by balancing the valid action distribution in every dialog state, $S_t$. Specifically, for each dialog turn $t$ with state-action pair $(S_t, A_t)$, we incorporate other valid system actions under the state $S_t$, i.e. $A_{t'}, t'\neq t$ with $S_{t'}=S_{t}$, as additional training data for turn $t$. The new objective function is: 
	\begin{equation}
	\mathcal{L}_{aug}=\sum_{t\in\mathcal{D}}\sum_{A_{t'}\in\mathcal{V^*}(S_t)}\log P(A_{t'}|S_t)
	\label{aug_loss}
	\end{equation}
	where $\mathcal{V^*}(S_t)\subseteq\mathcal{V}(S_t)$ is a subset of the valid action set $\mathcal{V}(S_t)$ of dialog state $S_i$. If we simply choose $\mathcal{V^*}(S_t)=\mathcal{V}(S_t)$, as every $P(A|S_t)$ is optimized by exactly the same number of each valid system action corresponding to state $S_t$, the overall conditional probability $P(A|S)$ is optimized on a balanced set of dialog actions. ???

	Our data augmentation framework over-samples training data to handle the unbalanced data problem. We choose over-sampling instead of under-sampling to make sure the dialog model can learn from all available dialogs. In practice, we can choose different $\mathcal{V^*}(S_t)$ to achieve different level of action diversity. For example, we find that for some system actions such as recommending a hotel name, a combination of other slots such as ``\textit{price}", ``\textit{stars}", ``\textit{parking}", ``\textit{wifi}" etc are often informed together as additional information, which makes the number of ``recommend" actions exponentially larger. However, they are semantically similar to each other. To avoid over-sampling of these actions, we group valid system actions with the same dialog act type together and uniformly sample from each group to form  $\mathcal{V^*}(S_t)$. This trick improve the learning efficiency and achieves a higher action diversity over different action types. In our experiments, we find some system actions are labeled incorrectly in MultiWOZ, which makes many dialog states only have one corresponding valid system action. To address this problem, we sample $\min(K, |G|)$ actions in each group, where $K>1$ is the predefined sample number and $|G|$ is the group size. This setting mitigates the influence of those unexpected single action groups but maintains the ability to learn from real single group action groups, e.g. rare cases in the dataset. Because large $K$ has a negative influence on the act-level balance over small groups. We empirically set $K=3$, as it yields the best experimental results.

	As a general learning framework, MADA is applicable to any task-oriented dialog system model that takes system actions as supervision, because without the system action annotation, we would not be able to obtain state-action mappings.  
	Our framework is suitable for all types of tasks as well. We choose the most challenging multi-domain task-oriented dialog corpus, MultiWOZ \cite{budzianowski2018multiwoz} to validate our framework's performance. We also designed a model, Domain Aware Multi-Decoder Network to take the full advantage of our data augmentation framework.

	\section{Domain Aware Multi-Decoder Network}
\begin{figure}[t!]
		\centering
		\includegraphics[width=1.0\columnwidth]{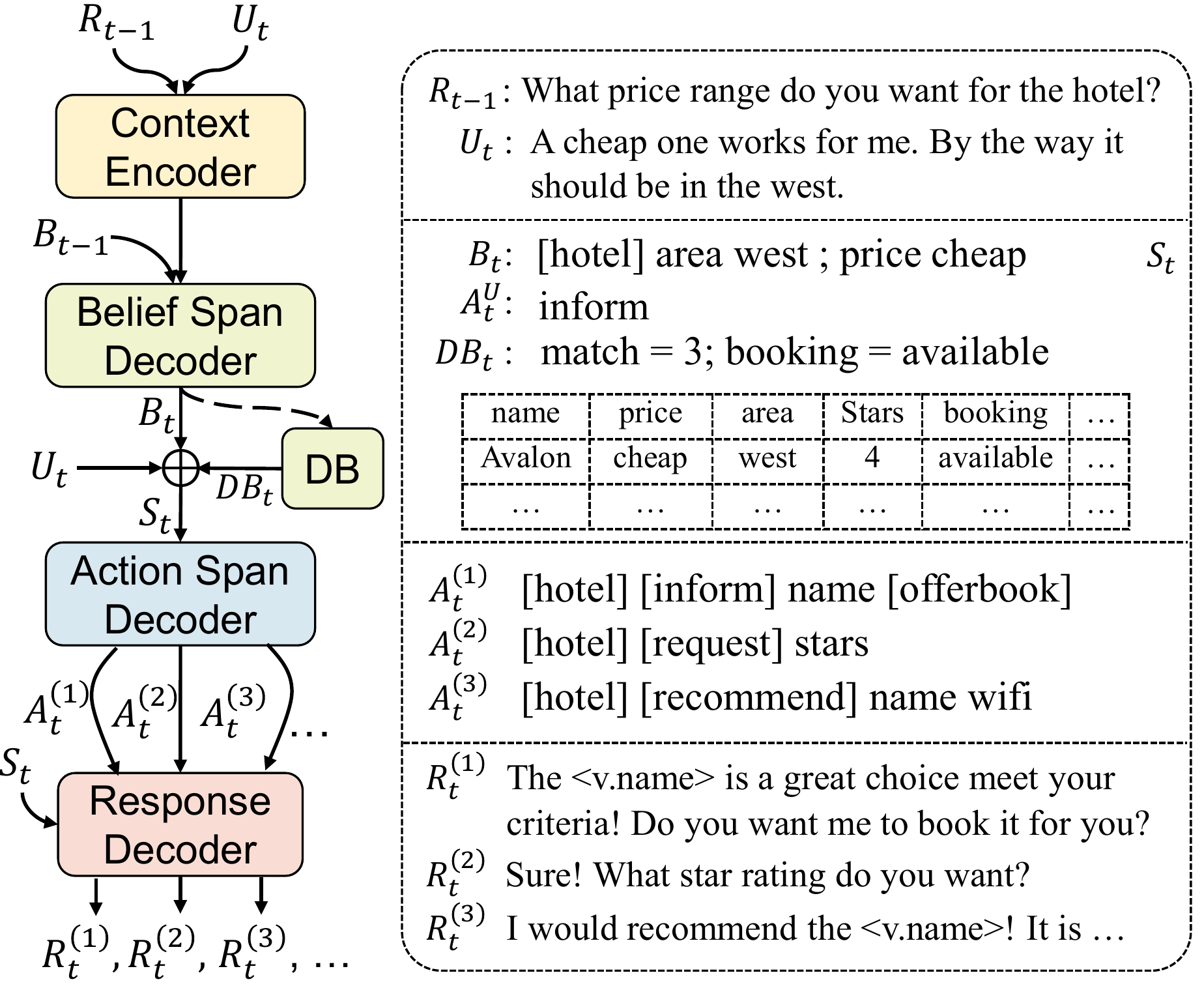} 
		\caption{The overview of Domain Aware Multi-Decoder (DAMD) network. The left figure shows the information flow among all modules. The explicit inputs and outputs of each module are described on the right.}
		\label{model}
	\end{figure}

	We propose a Domain Aware Multi-Decoder (DAMD) network, an end-to-end model designed to handle the multi-domain response generation problem through leveraging the proposed multi-action data augmentation framework. Fig.\ref{model} shows an overview of the proposed model. There are one encoder that encodes dialog context and three decoders that decodes belief span, system action and system response respectively. 
	
	\subsection{Domain-Adaptive Delexicalization}
	We first perform delexicalization to pre-process dialog utterances to reduce surface form language variability.
	Similar to \citeauthor{wen2017network} \shortcite{wen2017network}, we generate delexicalized responses with placeholders for specific slot values (see examples in Fig.\ref{model}), which can be filled according to database search results afterwards. However, we find that there is a drawback in the current multi-domain delexicalization scheme \cite{budzianowski2018multiwoz,chen2019semantically}. Previous methods only delexicalize the same slots in different dialog domains such as \textit{phone}, \textit{address}, \textit{name} etc as different tokens, e.g. \textit{\textless restaurant.phone\textgreater} and \textit{\textless hotel.phone\textgreater}, which adds extra burdens for the system to generate these critical tokens during task completion. We propose an adaptive delexicalization scheme using one token to represent the same slot name such as \textit{\textless v.phone\textgreater} in different dialog domains. Therefore the expressions in all relevant domains can be used to learn to generate the delexicalized value token. Since our model is domain-aware, the active domain is automatically updated based on dialog state. Therefore, there is no ambiguity in response generation process.\\

	\subsection{Belief Span Decoder}
	After data preprocessing, the model first learn to  decode belief span.
	The belief span $B_t$ of turn $t$ is updated based on the previous belief span $B_{t-1}$, previous system response $R_{t-1}$ and the current user utterance $U_t$ through a sequence to sequence fashion:
	\begin{equation}
	B_t={\rm seq2seq}(R_{t-1},U_t,B_{t-1})
	\end{equation}
	where the context vectors obtained by attention mechanism from the three sequences are concatenated to calculate the copy score. See \citeauthor{lei2018sequicity} \shortcite{lei2018sequicity} for more details. The copy mechanism is used to copy slot names, new slot values from utterances and unchanged parts of the previous belief span. Note that the entire history utterances are not used as the context information, since all the information is already contained and summarized in belief span. But the previous response is required, since the user may have some ellipsis in current utterance that refers to some slot values offered by system in the previous turn. The cross entropy between the generated and the ground truth belief spans are used as the loss of the belief span decoder. 
	
	In multi-domain dialog tasks, simply remembering the slot values instead of its dialog domain can lead to confusion. For example a time value can be either a reservation time in the restaurant domain or an arrival/leaving time for taxi booking. Therefore, we extend the belief span by decoding additional domain and slot tokens to address this ambiguity. An example multi-domain dialog state looks like ``\textit{{[restaurant]} name Curry Garden time 18:00 {[taxi]} leave 20:00 destination Kings Street}". The active dialog domain can automatically be determined by selecting the domain that recently changed semantic slot value. 
	
	For search results $DB_t$, we use an one-hot vector to indicate the number of matched entities and whether the booking is available or not following \citeauthor{budzianowski2018multiwoz} \shortcite{budzianowski2018multiwoz}.

	\subsection{System Action Span Decoder}
	The system action span decoder enables DAMD to utilize the multi-action data augmentation framework. We represent the system action as a sequence of tokens in the order of domains, acts and slots, as shown in the third text box in Fig. \ref{model}. 
	\begin{equation}
	A_t={\rm seq2seq}(U_t, B_t, DB_t)
	\end{equation}
	where the database search results are concatenated with the hidden state of the utterance and the belief states.
	We use the method described in the augmentation framework to enrich the training data. Specifically, for each system utterance in training, we find its dialog state based on annotation, which includes its dialog domain, belief state, database search result and its dialog act. Then we find this state's appropriate system action based on the previously learned one state to many system actions obtained in our data augmentation framework. The possible state-action pairs are used to enlarge the training set.

	In testing, our action decoder naturally has the ability to generate different system actions. Traditional beam search suffers from a diversity problem that the decoder tends to generate sequences with the same root 
	\footnote{For example, the model is more likely to generate ``\textit{[hotel] inform name price}" together with ``\textit{[hotel] inform name}" than ``\textit{[hotel] recommend name}" by standard beam search algorithm.}
	\cite{finkel2006solving,li2016simple}, 
	we address the issue by diversity promoting decoding techniques such as the diverse beam search \cite{li2016simple}, top-k sampling \cite{fan2018hierarchical} and nucleus sampling \cite{holtzman2019curious} to further introduce dialog policies diversity.
	
	\subsection{System Response Decoder}
	The final step is to generate response based on the dialog state and system action, which can be formulated as:
	\begin{equation}
	R_t={\rm seq2seq}(A_t, U_t, B_t, DB_t)
	\end{equation}
	where the hidden states of the belief span decoder and the action span decoder are used as $B_t$ and $A_t$. 
	Previous response decoder methods only base on system dialog act to decode sentences.  
	Our model is trained in an end-to-end manner, where all three decoders' loss are summed together and optimized jointly. 
	During evaluation, different responses are generated based on different system actions.

	\section{Dataset}
	We evaluate our proposed framework and model on the MultiWOZ dataset \cite{budzianowski2018multiwoz}. It is a large-scale human-human task-oriented dialog dataset collected via the Wizard-of-Oz framework where one participant plays the role of the system. 
	MultiWOZ contains conversations between a tourist and a clerk at an information center, which consists of seven domains including \textit{hotel}, \textit{restaurant}, \textit{attraction}, \textit{train}, \textit{taxi}, \textit{hospital} and \textit{police}, and an additional domain \textit{general} for some general acts such as greeting or goodbye. Each dialog in the dataset covers one to three domains, and multiple different domains might be mentioned in a single turn sometimes. Refer to \citeauthor{budzianowski2018multiwoz} (\citeyear{budzianowski2018multiwoz}) for statistics. Due to the multi-domain setting, complex ontology
	and flexible human expressions, developing dialog systems on MultiWOZ is extremely challenging.
	

	\section{Experimental Settings}
	
	\subsubsection{Pre-processing}
	
	The dataset is pre-processed through the proposed domain-adaptive delexicalization scheme as described in the previous section. The original belief state labels and system action labels are converted to the span form to train our domain-aware multi-decoder network model. The user action labels are adopted from the automatic annotations proposed by \citeauthor{lee2019convlab} (\citeyear{lee2019convlab}).

	\subsubsection{Automatic Evaluation Metrics}
	We focus on the context-to-response generation task proposed for MultiWOZ \cite{budzianowski2018multiwoz} and follow their automatic evaluation metrics. There are four automatic metrics to evaluate the response quality - if the system provides an correct entity (\textbf{inform rate}), answers all the requested information (\textbf{success rate}), is fluent \textbf{BLEU} \cite{papineni2002bleu} and a combined score \textbf{combined score} computed via $(Inform + Success)\times0.5+BLEU$ as an overall quality measure suggested in Mehri et al. \shortcite{mehri2019structured}. Since our goal is to learn diversed valid actions, we introduce two additional metrics to measure the action diversity: the number of unique type of dialog acts (\textbf{act number}) and slots (\textbf{slot number}) in all generated system actions in each dialog turn. In all of our experiments we report the average score of each metric over 5 runs.

	\begin{table}[t!]
		\small
		\centering
		\begin{tabular}{l|cc|cc}
			\hline
			\multirow{2}*{Model \& Decoding Scheme}  &\multicolumn{2}{c|}{Act \#} &\multicolumn{2}{c}{Slot \#} \\
			& w/o&w/&w/o&w/\\
			\hline
			\hline
			\multicolumn{5}{c}{Single-Action Baselines}\\
			\hline
			DAMD + greedy & \textbf{1.00} & \textbf{1.00}    &1.95 &\textbf{2.51}   \\
			HDSA + fixed threshold  & \textbf{1.00} & \textbf{1.00}    &2.07&\textbf{2.40}    \\
			\hline
			\hline
			\multicolumn{5}{c}{5-Action Generation}\\
			\hline
			DAMD + beam search & 2.67&\textbf{2.87}  &3.36&\textbf{4.39} \\
			DAMD + diverse beam search &2.68&\textbf{2.88}   &3.41&\textbf{4.50}  \\
			DAMD + top-k sampling &3.08&\textbf{3.43}  &3.61&\textbf{4.91} \\
			DAMD + top-p sampling &3.08&\textbf{3.40}  &3.79&\textbf{5.20} \\
			HDSA + sampled threshold& 1.32&\textbf{1.50}  &3.08&\textbf{3.31} \\ 
			\hline
			\hline
			
			\multicolumn{5}{c}{10-Action Generation} \\
			\hline
			DAMD + beam search & 3.06&\textbf{3.39}   &4.06&\textbf{5.29}  \\
			DAMD + diverse beam search &3.05&\textbf{3.39}   &4.05&\textbf{5.31}    \\
			DAMD + top-k sampling & 3.59&\textbf{4.12}   &4.21&\textbf{5.77}   \\
			DAMD + top-p sampling &3.53&\textbf{4.02}  &4.41&\textbf{6.17}   \\
			HDSA + sampled threshold  &1.54&\textbf{1.83}  &3.42&\textbf{3.92}\\
			\hline
		\end{tabular}
		\caption{Multi-action evaluation results. The ``w" and ``w/o" column denote with and without data augmentation respectively, and the better score between them is in bold. We report the average performance over 5 runs. }
		\label{multi} 
	\end{table}
	
	\begin{table*}[t!]
		\small
		\centering
		\begin{tabular}{l|c|cc|cccc}
			\hline
			\multicolumn{1}{c|}{\multirow{2}*{Model}} & Belief State & \multicolumn{2}{c|}{System Action} & \multirow{1}*{Inform} & \multirow{1}*{Success} & \multirow{2}*{BLEU} & Combined \\
			& Type & Type & Form & (\%)& (\%) &  & Score\\
			
			\hline
			1. Seq2Seq + Attention \cite{budzianowski2018multiwoz} & oracle & -&- & 71.3 & 61.0 & \textbf{18.9} &  85.1 \\
			2. Seq2Seq + Copy    & oracle &-&-& 86.2 & \textbf{72.0} & 15.7 &  94.8 \\
			3. MD-Sequicity  & oracle &-&-  & \textbf{86.6} & 71.6 & 16.8 &  \textbf{95.9} \\
			\hline
			4. SFN + RL (Mehri et al. \citeyear{mehri2019structured}) & oracle & generated & one-hot   & 82.7 & 72.1 & 16.3 &  93.7 \\
			5. HDSA \cite{chen2019semantically} & oracle & generated & graph & 82.9 & 68.9 & \textbf{23.6} &  99.5 \\
			6. DAMD  & oracle & generated & span   & \textbf{89.5} & 75.8 & 18.3 &  100.9 \\
			7. DAMD + multi-action data augmentation  & oracle & generated & span   & 89.2 & \textbf{77.9} & 18.6 &  \textbf{102.2} \\
			\hline
			8. SFN + RL (Mehri et al. \citeyear{mehri2019structured}) & oracle & oracle & one-hot   & - & - & 29.0 &  106.0 \\
			9. HDSA  \cite{chen2019semantically} & oracle & oracle & graph  & 87.9 & 78.0 & \textbf{30.4} &  113.4 \\
			
			10. DAMD + multi-action data augmentation   & oracle & oracle & span & \textbf{95.4} & \textbf{87.2} & 27.3 &  \textbf{118.5 }\\
			\hline
			11. SFN + RL (Mehri et al. \citeyear{mehri2019structured})  & generated & generated & one-hot  & 73.8 & 58.6 & \textbf{16.9} &  83.0 \\
			12. DAMD + multi-action data augmentation & generated & generated & span  & \textbf{76.3} & \textbf{60.4} & 16.6 &  \textbf{85.0} \\
			\hline
		\end{tabular}
		
		\caption{Comparison of response generation results on MultiWOZ. The oracle/generated denotes either using ground truth or generated results. The results are grouped according to whether and how system action is modeled. }
		\label{single} 
	\end{table*}

	\subsection{Baselines and Model Variations}
	We compare several model variations of our domain aware multi-decoder (DAMD) network  together with other base-
	lines on MultiWOZ. 
	\begin{itemize}
		\item 
		Seq2Seq + Attention \cite{budzianowski2018multiwoz}: a basic seq2seq model with attention (Bahdanau et al. \citeyear{bahdanau2014neural}).
		\item 
		Seq2Seq + Copy: a simplified version of DAMD where the belief and action span decoders are removed, which is equivalent to the copy-based seq2seq model \cite{gu2016incorporating}.
		\item 
		MD-Sequicity: a simplified version of DAMD with the action span decoder removed. We call it MD-Sequicity since it only extends the belief span to support multi-domain belief tracking comparing to the original Sequicity model \cite{lei2018sequicity}.
		\item 
		SFN + RL (Mehri et al. \citeyear{mehri2019structured}): a seq2seq network comprised of several pre-trained dialog modules which are connected through hidden states. Reinforcement fine tuning is used additionally to train the model. SFN is similar to our model in the spirit of modeling belief state and system action jointly in an end-to-end manner, but they use binary vectors for state and action modeling and do not take advantage of copying mechanism.
		\item
		HDSA: a hierarchical disentangled self-attention network \cite{chen2019semantically}. A BERT-based \cite{devlin2018bert} action predictor is used to predict system actions in HDSA. Since the original multi-label classification with a fixed active threshold is not able to generate multiple actions, we alternatively samples a threshold for each dimension of the action vector independently. The actions are used to control the structure of a self-attention network afterwards for response generation, which is trained separately with the action predictor. 
		\item
		DAMD: our proposed domain aware multi-decoder network. The belief state, system action and response are generated in a seq2seq manner in DAMD. We use greedy decoding for all single-sequence decoding process. When decoding multiple actions, we leverage the standard beam search algorithm and several diversity-promoted decoding schemes :  \\
		(1) the diverse beam search (Li et al. \citeyear{li2016simple}) which adds a penalty term to intra-sibling sequences thus favors choosing hypotheses from diverse parents. \\
		(2) the top-$k$ sampling algorithm (Fan et al. \citeyear{fan2018hierarchical}) which samples the next word from the $k$ most probable choices according to vocabulary distribution. \\
		(3) the top-$p$ sampling algorithm \cite{holtzman2019curious} which samples from the set of top possible words where their summed probability reaches a fixed value $p$.
	\end{itemize}
	
	\subsubsection{Parameter Setting}
	In our implementation of DAMD, we use a one-layer bi-directional GRU with hidden size of 100 as encoder and three standard GRUs with the same hidden size as decoders. The embedding size, vocabulary size and batch size are 50, 3,000 and 128 respectively. The combined score on development set is used as the validation check metric. We use the Adam optimizer with a initial learning rate of 0.005. The learning rate decays by half every 3 epochs if no improvement is observed on development set. Training early stops when no improvement is observed on development set for 5 epochs. For multi-action decoding, the beam size and sampling number $k$ are the same as action number, which is 5 or 10 in our experiments. We use 0.2 as the diverse beam search penalty and  $p=0.9$ for top-$p$ sampling. The fixed active threshold for HDSA is 0.4, and the sampling range is $\left[0.3, 0.5\right]$ in multi-action experiments. All of the hyperparameters are selected through grid search. The code is available here\footnote{https://gitlab.com/ucdavisnlp/damd-multiwoz}.

	\section{Results and Analysis}
	We first evaluate whether our data augmentation framework efficiently improves dialog policy diversity. We conduct experiments of 5-action and 10-action generation, where different model variations with and without utilizing the proposed multi-action data augmentation framework are compared. The results are shown in Table \ref{multi}. After applying our data augmentation, both the action and slot diversity are improved consistently, which indicates that our data augmentation framework is applicable to different models. Top-$k$ sampling achieves the highest act-level diversity, where there are 3.43 unique dialog acts on average in five generated actions. HDSA has the worse performance and benefits less from data augmentation comparing to our proposed domain-aware multi-decoder network, because HDSA does not decode its dialog act but perform multi-label classification. While the appropriateness of multiple actions is hard to judge by automatic evaluation \cite{liu2016not}, we leave it for human evaluation, where we also take a further step to directly evaluate the corresponding responses. 
	
	We evaluate our domain-aware multi-decoder (DAMD) network on the context-to-response generation task based on MultiWOZ. Each model generates one response for fair comparison. Experiments with ground truth belief state feed the oracle belief state as input and database search condition. Specifically in DAMD, we feed the oracle token at each decoding step of belief span to produce the oracle hidden states as input of subsequent modules. 
	Results are shown in Table \ref{single}. 
	The first group shows that after applying our domain-adaptive delexcalization and domain-aware belief span modeling, the task completion ability of seq2seq models becomes better. The relative lower BLEU score is potentially due to that task-promoting structures (e.g. copy) make the model focus less on learning the language surface. Our DAMD model significantly outperforms other models with different system action forms in terms of inform and success rates, which shows the superiority of our action span. While we find applying our data augmentation achieves a limited improvement on combined score (6 vs 7), which suggests learning from a balanced state-action training data can improve the robustness of model but the benefit of learning diverse policy for single response generation is hard to evaluate. Moreover, if a model has access to ground truth system action, the model further improves its task performance. Finally, we find conditioned on generated belief state greatly harm the response quality, due to the error propagation from previous decoders to the final response decoder. Note that HDSA cannot track belief state thus has no results here. 
	\begin{table}[t!]
		\scriptsize
		\centering
		\resizebox{.99\columnwidth}{!}{ 
			\begin{tabular}{l|l}
				
				
				\hline
				
				\multicolumn{2}{l}{\textbf{SYS}: {\fontsize{6.4pt}{6.4pt} \selectfont I also have 3 pricing options and amenity options. Could you give me some direction?}}\\
				\multicolumn{2}{l}{\textbf{USER}: {\fontsize{6.5pt}{6.5pt} \selectfont Sure. 4 star, nothing but the best, free wifi moderately priced and free parking too.}} \\
				\multicolumn{2}{l}{\textbf{STATE}:\ {\fontsize{6.5pt}{6.5pt}  [hotel] parking yes; pricerange moderate; stars 4; internet yes; \  DB-match:\textbf{9}}} \\
				
				\hline
				\multicolumn{1}{c|}{Generated Actions w/o MADA} &\multicolumn{1}{c}{Generated Actions w/ MADA}\\
				\hline
				{\textbf{[inform] area choice price}}& [inform] choice [request] area\\
				{[inform] area price choice type}& [inform] name internet parking area [offerbook] \\
				{[inform] choice [request] area}& [recommend] name [inform] choice\\
				{[inform] choice [request] area price}& \textbf{[recommend] name [offerbook]} \\
				{[inform] choice type [request] area}&  [recommend] name [inform] choice [offerbook] \\
				\hline
				\textbf{SYS}: There are 9 moderate places  &\textbf{SYS}: I would suggest acorn guest house.   \\
				~~~~~~~~~ in the north.& ~~~~~~~~ Would you like me to book you a room? \\
				\hline
				\multicolumn{2}{l}{\textbf{GT Action}: [recommend] price name [offerbook]}\\
				\multicolumn{2}{l}{\textbf{GT SYS}: May I recommend acorn guest house? It is moderate and fits all your}\\
				\multicolumn{2}{l}{~~~~~~~~~~~~~~~ criteria. Would you like me to reserve you any rooms?}\\
				
				\hline
		\end{tabular}}
		\caption{Our model's example generation responses with and without data augmentation. GT denotes the ground truth. The generated action candidate closest to the ground truth action is marked in bold.}
		\label{case} 
	\end{table}

	\begin{table}[t!]
		\scriptsize
		\centering
		\resizebox{.99\columnwidth}{!}{
			\begin{tabular}{c|l|l}
				\hline
				\multirow{9}*{\shortstack{Policy \\ Error}} 
				&{\fontsize{6.4pt}{6.4pt} \selectfont\textbf{USER}: I will be travelling from Cambridge} 
				&{\fontsize{6.4pt}{6.4pt} \selectfont\textbf{USER}: Yes, i would like a reservation.} 
				\\
				&{\fontsize{6.4pt}{6.4pt} \selectfont actually and going to London Kings Cross.}
				& 
				\\
				
				&\textbf{GT Action}: [request] day& \textbf{GT Action}: [request] people \\
				\cline{2-3}
				&\textbf{Generated Actions}: &\textbf{Generated Actions}:\\
				&[request] leave & [offerbook]\\
				&[request] leave arrive & [offerbook] [general] [reqmore]\\ 
				&[inform] choice [request] leave & [inform] food area name [offerbook]\\
				&[inform] choice [request] leave arrive & [inform] address name [offerbook]\\
				&[inform] id arrive destination [offerbook] & [inform] food area name [reqmore]\\
				\hline
				\multirow{3}*{\shortstack{NLG \\ Error}}
				& \multicolumn{2}{l}{\textbf{Action}: [recommend] name [inform] postcode address [reqmore]}\\
				& \multicolumn{2}{l}{\textbf{SYS}: I would suggest the\ \textless v.name\textgreater. The postcode is \ \textless v.postcode\textgreater.} \\
				& \multicolumn{2}{l}{~~~~~~~~ Is there anything else I can help you with today?}\\
				\hline
			\end{tabular}
		}
		\caption{Examples of errors made by DAMD. GT denotes ground truth.}
		\label{error} 
	\end{table}


	\subsection{Case Study and Error Analysis}
	Table \ref{case} shows an example where learning policy diversity is beneficial for task completion. Since there are still nine hotels which fit the user's requirement, a common policy should be requesting a slot (e.g. area located) to further reduce database search range. However, the dialogs are carried out by a large number of different crowd workers. Some workers may choose to make a direct recommendation instead. This less frequency seen policy is difficult for the system to capture without a balanced data set, as the model tend to generate only the majority \textit{request} actions. After applying data augmentation, the \textit{recommend} actions are also captured as a valid action.
	
	Although better than models trained on unbalanced state-action dataset, our model still makes several types of errors shown in Table \ref{error}. The example on the left shows the model makes an error on the slot type. This is because our data augmentation method mainly focuses on improving the act-level policy diversity and the slot-level diversity is ignored. The example on the right shows an error where our system failed to collect enough information, such as number of people, before offering to make a restaurant reservation. This suggests that more prior task knowledge should be injected in the dialog model to address such issue. Moreover, besides policy level errors, there is also errors in the response generation process. In the bottom example shown in Table 5, the address information of the restaurant is missing. Such error might be caused by the generation model forgetting the distant information when the conditioned action span is too long.

	\subsection{Human Evaluation}
	
	Automatic metrics only validate system’s performance on one single dimension at a time. While human can provide an ultimate holistic evaluation. We conduct human evaluation to show that learning a balanced dialog policy can eventually improve the dialog system responding quality, in terms of higher appropriateness of individual responses and higher diversity among multiple responses. 
	
	In our experiments, \textbf{appropriateness} is scored on a Likert scale of 1-3 which denotes \textit{invalid}, \textit{ok} and \textit{good} respectively, for each generated response. \textbf{Diversity} is scored on a Likert scale of 1-5 for all of the responses (we generate 5 responses for each model in our experiments). We suggest the judges to score according to the number of different policies in responses. We evaluate three models: DAMD without data augmentation, DAMD with data augmentation and HDSA with data augmentation. The top-$k$ sampling is selected as our decoding methods since it achieves highest action diversity as shown in Table \ref{multi}. We sample one hundred dialog turns and the 15 responses (five responses for each model) of each turn are scored by three judges given the dialog history. 
	
	The results are shown in Table \ref{human}. We report the average value of diversity and appropriateness, and the percentage of responses scored for each appropriateness level. With data augmentation, our model obtains a significant improvement in diversity score and achieves the best average appropriateness score as well. Due to the larger diversity, DAMD with augmentation is more likely to generate responses with better quality. However, the slightly increased invalid response percentage indicates that some invalid actions are also captured, which may due to that noisy state and action labels lead to wrong valid state-action set. We also observe our DAMD model outperforms HDSA in both diversity and appropriateness scores. This is mainly because our model considers the dialog domain information in a more effective manner and our model is able to leverage the state-action augmentation better by decoding system actions instead of performing classification. 
	In summary, the overall results suggest that our framework can effectively improve the ability of dialog systems to generate appropriate responses with different dialog policies.  
	
	\begin{table}[h]
		\small
		\centering
		\resizebox{1.0\columnwidth}{!}{
			\begin{tabular}{l|c|c|ccc}
				\hline
				Model & Diversity & App & Good\% & OK\% & Invalid\% \\
				\hline
				DAMD          &3.12   &2.50   &56.5\%   &\textbf{37.4\%}  &6.1\% \\
				DAMD (+)   &\textbf{3.65}   &\textbf{2.53}   &\textbf{63.0}\%   &27.1\%  & 9.9\% \\
				HDSA (+)   &2.14   &2.47   &57.5\%   &32.5\%  &\textbf{10.0\%} \\
				
				\hline
		\end{tabular}}
		\caption{Human evaluation results. Models with data augmentation are noted as (+). App denotes the average appropriateness score.  }
		\label{human} 
	\end{table}

	\section{Conclusion}
	We focus on generating appropriate responses with higher diversity in task-oriented dialog systems, by learning a diversified dialog policy through considering the one-to-many dialog property. Specifically, we propose the Multi-Action Data Augmentation (MADA) framework to enable dialog models to learn a more balanced state-to-action mapping. Our framework generalizes to all dialog tasks with belief state and system action annotated. We also propose a new 
	domain aware multi-decoder (DAMD) model to leverage the proposed data augmentation framework. DAMD learns a more diverse state-to-action policy which not only achieves the state-of-the-art task success rate on the challenging MultiWOZ dataset, but also generates a set of responses that are both appropriate and diverse. In the future we plan to apply our method to help the modeling of diverse user behaviors.


	\fontsize{9.0pt}{10.0pt} \selectfont
	\bibliography{ref}
	\bibliographystyle{aaai}
	
\end{document}